# Innovative Framework for Early Estimation of Mental Disorder Scores to Enable Timely Interventions

1st Himanshi Singh
*Dept. of Information Technology*
*IIIT Allahabad*
Prayagraj, Uttar Pradesh, India
prf.himanshi@iiita.ac.in

2nd Sadhana Tiwari
*Dept. of Information Technology*
*IIIT Allahabad*
Prayagraj, Uttar Pradesh, India
rsi2018507@iiita.ac.in

3rd Sonali Agarwal
*Dept. of Information Technology*
*IIIT Allahabad*
Prayagraj, Uttar Pradesh, India
sonali@iiita.ac.in

4th Ritesh Chandra
*Dept. of Information Technology*
*IIIT Allahabad*
Prayagraj, Uttar Pradesh, India
rsi2022001@iiita.ac.in

5th Sanjay Kumar Sonbhadra
*Dept. of Computer Science and Engg.*
*Siksha 'O' Anusandhan University*
Bhubaneswar, Odisha, India
sksonbhadra@gmail.com

6th Vrijendra Singh
*Dept. of Information Technology*
*IIIT Allahabad*
Prayagraj, Uttar Pradesh, India
vrij@iiita.ac.in

*Abstract*—Individuals' general well-being is greatly impacted by mental health conditions including depression and Post-Traumatic Stress Disorder (PTSD), underscoring the importance of early detection and precise diagnosis in order to facilitate prompt clinical intervention. An advanced multimodal deep learning system for the automated classification of PTSD and depression is presented in this paper. Utilizing textual and audio data from clinical interview datasets, the method combines features taken from both modalities by combining the architectures of LSTM (Long Short-Term Memory) and BiLSTM (Bidirectional Long Short-Term Memory). Although text features focus on speech's semantic and grammatical components; audio features capture vocal traits including rhythm, tone, and pitch. This combination of modalities enhances the model's capacity to identify minute patterns connected to mental health conditions. Using test datasets, the proposed method achieves classification accuracies of 92% for depression and 93% for PTSD, outperforming traditional unimodal approaches and demonstrating its accuracy and robustness.

## I. INTRODUCTION

Mental health conditions such as depression and post-traumatic stress disorder (PTSD) impact millions of people worldwide, becoming major global health concerns. In addition to lowering people's quality of life, many illnesses have a significant negative impact on society and the economy. If not treated or recognized, mental health issues can lead to chronic diseases, decreased functioning, and even higher death rates. In under-resourced areas mental health issues are prevalent, even with advancements in clinical practice, traditional methods of diagnosing these disorders—such as psychological testing and in-person interviews—are still limited due to their subjective nature, resource-intensive nature, and reliance on the availability of qualified healthcare professionals.

The increasing need for prompt and precise diagnosis, along with the shortage of qualified mental health specialists in many places, has made the incorporation of automated systems in mental health treatment possible. New developments in deep learning (DL) and machine learning (ML) technology have shown a lot of promise in tackling these issues. By evaluating many data modalities, including text, audio, and even visual clues, these automated systems offer scalable, objective, and effective methods of recognizing mental health disorders, allowing for earlier and more individualized therapies. A supplementary tool for clinical decision-making, machine learning algorithms—in particular, deep neural networks—have shown the ability to identify subtle patterns in a variety of datasets that conventional approaches would overlook. We introduce an automated technique for detecting early PTSD and depression in this study. Our goal is to create a deep learning model that can detect mental signals. Use the Distress Analysis Interview Corpus-Wizard of Oz (DAIC-WOZ) collection to evaluate health conditions through written transcripts, audio recordings, and clinically verified depression ratings. The DAIC-WOZ dataset provides synchronized data from many modalities, providing a full model of psychological distress, including textual and voice cues.

This framework integrates textual and aural input to detect depression and PTSD using multimodal deep neural networks. Audio qualities like tone, pitch, and rhythm capture paralinguistic cues that indicate emotional state and psychological well-being. Text embeddings analyze spoken language for emotional content and cognitive distortions linked to mental health conditions.

We structure the document as follows: Section II provides a thorough analysis of the body of research, focusing on developments in mental health diagnosing techniques. Section III notes the shortcomings and difficulties of the existing methods. Section IV provides a detailed description of the study's structure and methods, along with an explanation of the proposed approach. Section V provides a comprehensive description of the dataset, detailing the preparation procedures, data sources, and feature extraction procedures. While Section VII wraps up the study and makes recommendations for possible future research areas, Section VI examines the outcomes and important discoveries.

## II. LITERATURE REVIEW

Recent developments in emotional computing have led to the development of automated depression detection systems. Jo et al. used Convolutional Neural Networks (CNN) and Bidirectional Long Short-Term Memory (Bi-LSTM) to present a four-stream model that combines text and audio data for diagnosis. Jo et al. retrieved and examined audio properties like MFCC, GTCC, and Mel spectrograms, and numerically encoded the text data. Each stream's ensemble of softmax outputs served as the basis for the model's diagnostic. Using the Extended Distress Analysis Interview Corpus (E-DAIC-WOZ) and other datasets, tests showed that three-stream models performed 10.7% to 11.9% better than two-stream models [1].

Jung et al. created HiQuE (Hierarchical Question Embedding network) to improve the automatic detection of depression by looking at the hierarchical structure of clinical interview questions. HiQuE enhances assessment by linking main and follow-up questions, unlike previous methods that just analyzed interview data. The technique utilizes mutual knowledge across many modalities to determine the significance of each question. A lot of testing on the DAIC-WOZ dataset showed that HiQuE is better than current multimodal models at identifying emotions and depression. This indicates improved early diagnostic and treatment applications [2].

Ding et al. developed IntervoxNet, a sophisticated computer-aided detection approach to enhance depression diagnosis using interview audio analysis. The system employs two methods: the Audio Mel-Spectrogram Transformer (AMST) for audio analysis and the BERT-CNN model for text processing. IntervoxNet outperformed existing state-of-the-art approaches on the DAIC-WOZ dataset with an F1 score of 0.90, recall of 0.92, precision of 0.88, and accuracy of 0.86. The findings suggest its potential as an effective tool for clinical automated depression screening [3].

Sadeghi et al. developed an automated depression severity prediction algorithm using the E-DAIC dataset. The work uses large language models (LLMs) to extract depression-related signals from interview transcripts using PHQ-8 scores as training data. The study also constructs a multimodal prediction model using text and video frame face data. The study evaluates facial, text-based, and combined techniques. Adding voice quality ratings to text data yields the best results, with a mean absolute error of 2.85 and a root mean square error of 4.02. These findings show that text-only models function and that multimodal techniques may improve automated depression recognition. [4].

Chen et al. used text, photos, and social behavior variables from Weibo to create the Text Picture Sentiment Auxiliary (TPSA) model for depression identification. They employ tree-LSTM and Bi-GRU models to capture text dependencies and a text extension strategy to handle text length variability. They extract emotional elements from photos using OCR technology and an emotion vocabulary.His model effectively detects depression with an accuracy of 0.987 and recall of 0.97 after processing data using Fusion Network Attention [5].

Kawade et al. examined the expanding area of speech emotion recognition (SER), highlighting the potential uses of this technology in human-computer interactions (HCI). Recent research has moved toward deep learning models, which eliminates the requirement for manual feature identification, but classic SER approaches still rely on handmade feature extraction. However, implementing deep learning algorithms on standalone processing boards is difficult due to their high-dimensional characteristics. The authors addressed the issue by presenting a multicore PYNQ-ZQ FPGA board-based deep learning SER system. With an accuracy of 85.33% and a lower processing time than traditional CPU-based solutions, this system highlighted the promise of FPGA for effective SER deployment [6].

Wang et al. developed a novel approach to diagnosing depression using speech acoustic features, addressing limitations of traditional methods.Wang et al. developed a 3D-CBHGA convolutional filter bank using highway networks and a bidirectional GRU with an attention mechanism to capture complex speech patterns associated with sadness.The approach enhances feature extraction and utilizes the GRU network's attention mechanism to recognize depression-related emotional signals in speech. The trial revealed that 3D-CBHGA significantly improved depression detection accuracy, suggesting its potential for more efficient and objective diagnosis [7].

By combining text, video, and audio data, Zhang et al. created the Audio, Video, and Text Fusion-Three Branch Network (AVTF-TBN) for detecting depression risk. The model extracts the characteristics of each modality and then aggregates them for predictive analysis. The model demonstrated the efficacy of multimodal sensor-based data in identifying depression risk by achieving an F1 score of 0.78, precision of 0.76, and recall of 0.81 using a dataset of emotion-elicitation tasks [8].

Mai et al. investigated the idea of brain-conditional multimodal synthesis, or AIGC-Brain, which generates several modalities including text, graphics, and audio by using brain signals as a guide. These brain signals are useful for multimodal synthesis because they show a One-to-Many link with many modalities and represent how the brain interprets outside information. The study explores the generative models and neuroimaging datasets that serve as the foundation for AIGC-Brain, offering a thorough taxonomy and talking about how it might improve brain-computer interface systems and help us better understand how humans perceive the world [9].

In order to overcome the drawbacks of conventional clinical interviews and self-report measures, Dibekliog˘lu et al. looked into the use of facial, head movement, and vocalization dynamics for diagnosing depression severity. They trained classifiers on a population sample with severe depressive illness, evaluated at regular intervals over a 21-week period, using these behavioral characteristics. Their results showed that voice prosody performed the least well in identifying the degree of depression, whereas facial movement dynamics offered the best accuracy, followed by head movement dynamics. Combining the dynamics of head and face movements yielded the highest results, demonstrating the potential of multimodal techniques for more precise and impartial depression identification [10]

In order to diagnose depression, Qureshi et al. presented a unique multitask learning attention-based deep neural network model that incorporates textual, visual, and audio input. By utilizing the advantages of both techniques, their method completes regression and classification tasks concurrently. Experiments on the Distress Analysis Interview Corpus (DAIC-WOZ) showed that multitask learning networks trained on both regression and classification worked better than single-task models. Furthermore, combining all modalities yielded the most accurate depression assessment, highlighting the advantages of a multimodal strategy in enhancing the accuracy of depression identification [11].

Lin et al. proposed an autonomous depression identification method using speech signals and linguistic information from patient interviews. Their They use a one-dimensional convolutional neural network (1D CNN) to look at voice, a bidirectional long short-term memory (BiLSTM) network with an attention layer to look at text, and a fully linked network that combines both to look at sadness. method demonstrated enhanced identification accuracy and attained state-of-the-art performance when tested on two publicly available datasets [12].

Zhang et al. proposed the hybrid deep learning model RoBERTa-BiLSTM to enhance depression identification through feature extraction from text sequences. Their method combines the advantages of both transformers and sequence models, acknowledging the difficulties caused by lengthy calculation durations in sequence models. The model reduces processing time without compromising accuracy by utilizing RoBERTa to construct word embeddings and BiLSTM to capture long-distance contextual semantics. [13].

## III. DETECTION OF MENTAL HEALTH DISORDERS: EXISTING GAPS

A number of obstacles prevent the detection of mental health issues from being used effectively. Many current systems are limited in their capacity to capture the complexity of psychological circumstances because they rely on unimodal methodologies, which only use text, visual, or aural input. Furthermore, the generalizability of models across various populations and demographics is diminished in the lack of sizable, varied, and balanced datasets, particularly for underrepresented groups. Since exact comprehension of speech context throughout time is necessary to detect minor emotional changes, temporal modeling continues to be a substantial issue. Furthermore, physiological signals like heart rate, skin reaction, and brainwave data are commonly overlooked, despite the fact that they have the ability to increase detection accuracy and provide a more complete knowledge of mental health issues. Handling sensitive personal data often disregards ethical and privacy considerations like permission and data protection, creating compliance and trust challenges. These models are less useful in emergency treatment because they lack real-time, deployable technologies to stimulate clinical interventions. The lack of standardized mental health diagnostic tools makes assessing and comparing interventions difficult.

## IV. METHODOLOGY

### A. Workflow

Figure 1 shows how the depression detection algorithm collects text and audio input. After processing these inputs, the model extracts key properties and analyzes them to detect depressive symptoms. The model may identify emotional patterns and reveal mental health by using an organized technique.

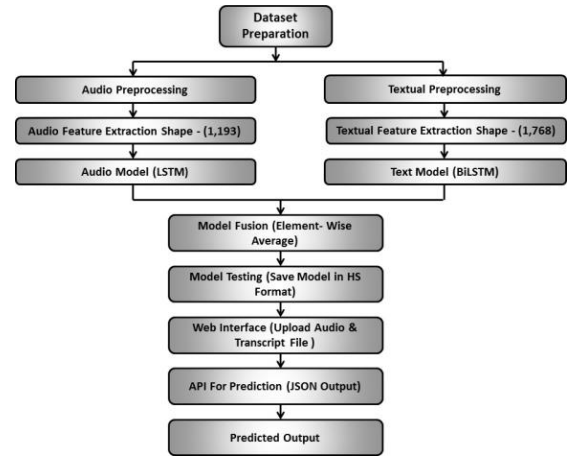

Fig. 1: Model Workflow

### B. Model for Detecting Depression

The depression detection model processes audio and text inputs smoothly using a dual-branch architecture. The model receives audio characteristics as vectors (1, 193) and text embeddings as vectors (1, 768). The text branch uses dropout layers for regularization, dense layers for feature modification, and a bidirectional LSTM layer to capture temporal dependencies in text embeddings. The audio branch works on sequential audio features at the same time, using an LSTM layer with dropout and thick layers to avoid overfitting and improve feature learning. Binary classification can predict sadness by averaging the outputs from these two branches and passing them through a thick layer with a sigmoid activation function. This integrated and modular architecture maximizes acoustic and semantic variables, improving model prediction.
.

## C. The LSTM Model

In order to capture dependencies, the Long Short-Term Memory (LSTM) model processes sequential data. LSTMs excel at voice recognition, time-series forecasting, and temporal analysis. An input, forget, and output memory method allows LSTMs to choose to keep, update, or delete data. LSTMs can learn from extended sequences because their structure solves the disappearing and expanding gradient difficulties with RNNs. Applications for sequential data use LSTMs.

Bidirectional LSTM (BiLSTM) models improve standard LSTMs by analyzing input sequences both forward and backward, allowing for the collection of both past and future contextual information. This is particularly helpful for jobs where comprehending the complete context of a sequence is crucial, such as sentiment analysis or machine translation. For a more thorough sequence representation, the model first runs input through a forward LSTM and then a reverse LSTM, integrating their results. BiLSTMs work well in situations where both past and future context are important, such as speech recognition and natural language processing. In order to ensure steady learning, they additionally employ gates to control information flow and solve the vanishing gradient issue.

*1) LSTM for Audio:* LSTM models can capture temporal patterns in audio sources, making them ideal for speech recognition, emotion detection, and audio categorization. For jobs that need data order, such as voice frames or acoustic characteristics, LSTMs can use context from prior frames by examining sequences step by step. In audio data, pitch, tone, and prosody fluctuate over time. LSTMs may describe these dependencies. This makes them ideal for dynamic audio signal analysis.

*2) BiLSTM for Text:* BiLSTMs (Bidirectional Long Short-Term Memory networks) perform well in text processing because they capture context from previous and subsequent words. Dual processing helps the model grasp word meaning for named entity identification, sentiment analysis, and machine translation. BiLSTMs increase long-range dependency and text representation by examining the sequence left-to-right and right-to-left. They are ideal for difficult linguistic problems that involve past and future context.

## D. PTSD Detection Model

The architecture of the depression detection model and the Post-Traumatic Stress Disorder (PTSD) detection model is comparable, using the same multimodal data fusion methodologies, data preparation methods, and input dimensions. For both models, this consistency guarantees effective processing and analysis. The PTSD model is unique in that it uses labels related to PTSD throughout training, allowing it to concentrate on identifying PTSD-specific patterns and signs. This approach allows the model to diagnose PTSD with high accuracy while maintaining interoperability with other mental health detection techniques.

ALGORITHM: INNOVATIVE FRAMEWORK FOR EARLY ESTIMATION OF MENTAL DISORDER SCORES

**Algorithm 1** Innovative Framework for Early Estimation of Mental Disorder Scores
1: **Step 1: Load Features**
2: Load text and audio features from the respective data sources.
3: Store features in separate arrays: `TextFeatures`, `AudioFeatures`.
4: Load binary classification labels into `Labels`.
5: **Step 2: Prepare Data**
6: Reshape `TextFeatures` and `AudioFeatures` for compatibility with LSTM input dimensions.
7: Split `TextFeatures`, `AudioFeatures`, and `Labels` into training and testing sets (80%-20%).
8: **Step 3: Define Model Structure**
9: Text branch: `TextInput` → `BiLSTM` → `Dropout` → `Dense`.
10: Audio branch: `AudioInput` → `LSTM` → `Dropout` → `Dense`.
11: Fuse outputs from both branches using element-wise average.
12: **Step 4: Compile Model**
13: Compile the model using Adam optimizer (learning rate: 0.001) and binary cross-entropy loss.
14: **Step 5: Train Model**
15: Train with batch size = 8, epochs = 8, validation split = 0.2.
16: **Step 6: Evaluate Model**
17: Evaluate the model on the test set and record performance metrics (e.g., accuracy, loss).
18: **Step 7: Predict and Classify**
19: Use the model to predict mental disorder scores for new text and audio features.
20: Compare scores against a predefined threshold:
21: **if** score > threshold **then**
22:     Trigger Early Intervention.
23: **else**
24:     Schedule Regular Monitoring.
25: **end if**

Algorithm 1 provides an example of the novel framework for early mental disorder score estimation

Preprocessing included lowering sample sizes for LSTM input after extracting and normalizing text and audio features from CSV files. We employed a bidirectional LSTM (64 units) for text and a standard LSTM (64 units) for audio. We divided the dataset 80/20 for training and testing, using 64-unit thick layers and a 0.3 dropout to prevent overfitting. We used a sigmoid output for binary classification, and employed element-wise averaging to achieve audio-text fusion. We trained the model for ten epochs using TensorFlow, Keras, Scikit-learn, Pandas, and Matplotlib tools, using binary cross-entropy loss, the Adam optimizer (learning rate 0.001), and a batch size of eight.

## V. DATASET OVERVIEW

### A. Description of the Dataset

The DAIC-WOZ (Distress Analysis Interview Corpus - Wizard of Oz), (https://dcapswoz.ict.usc.edu/wwwedaic/) [14] collection was created especially for mental health research, especially in the area of depression identification. A human operator controls the virtual interviewer's comments and interactions with the participants in a sequence of clinical interviews that are done in a Wizard-of-Oz-like setting. This dataset is perfect for machine learning and computational analysis in the context of mental health diagnostics since it includes a diverse variety of participant sessions, each of which provides rich multimodal data, including text, audio, and video.

### B. Components of the Dataset

The features used from text and audio data to predict PTSD and depression are shown in Table I, along with pertinent classification-related attributes.

TABLE I: Extracted Features from Audio and Text Data

| Feature Name | Shape | Description |
|---|---|---|
| **Audio Features(193)** | | |
| MFCCs | (13,) | Captures main audio spectral features. |
| Delta MFCCs | (13,) | Shows changes in MFCC values over time. |
| Delta2 MFCCs | (13,) | Represents acceleration of MFCC changes. |
| Chroma | (12,) | Energy across 12 musical pitch classes. |
| Mel Spectrogram | (128,) | Frequency intensity in the perceptual scale. |
| Contrast | (7,) | Measures spectral peaks and valleys. |
| Tonnetz | (6,) | Encodes harmonic and tonal information. |
| Pitch | (1,) | Dominant frequency in the audio signal. |
| **Text Features(768)** | | |
| BERT Embedding | (768,) | Contextual word representations capturing semantic and syntactic information of text. |

Audio Data: The DAIC-WOZ dataset's audio data is supplied in high-quality WAV format with a sampling rate of 16 kHz, guaranteeing sound recordings of clinical quality appropriate for precise feature extraction. With a duration of 20 to 60 minutes each, there is sufficient material for a thorough examination. Both acoustic and prosodic aspects are crucial for identifying possible indicators of sadness and PTSD, and this format and quality are perfect for extracting both.

Text Data: Time-stamped conversations that enable exact synchronization with the associated audio recordings are included in the CSV format of the textual transcripts. These transcripts provide crucial linguistic and contextual information for study by capturing the verbal content of every session. The structured text data, when paired with the audio, allows for a more thorough multimodal analysis, allowing sentiment analysis and natural language processing to provide important information about mental health issues.

### C. Feature Extraction

The figure 2 depicts the system's architecture, including how text and audio features are extracted and incorporated into models for PTSD and depression prediction. By highlighting the crucial phases from input processing to prediction output, it illustrates the multimodal method for precise classification of mental health disorders.

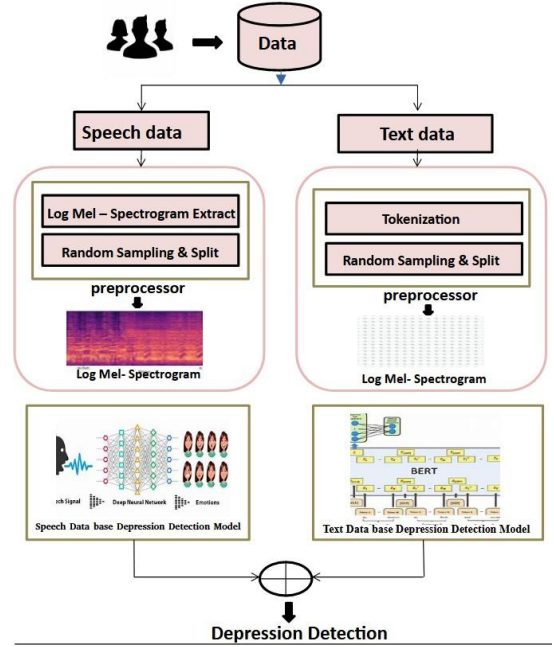

Fig. 2: Model Diagram Mental Disorders

The first step is to obtain audio recordings and the transcripts that go with them. After preprocessing the audio, which includes segmentation and silence removal, feature extraction methods such MFCCs, delta MFCCs, and spectral features are applied. Similarly, before extracting text features—often with the help of techniques like BERT embeddings—the transcripts undergo preprocessing through tokenization, lemmatization, and stopword removal. Then, possibly via concatenation or more complex methods, these collected text and audio features are combined. The combined features are then entered into machine learning models that have been trained and are especially tailored to predict sadness and PTSD.

*1) Overview of the System:* The Hadoop Distributed File System (HDFS) cluster is used in the feature extraction system

to facilitate parallel processing and effective distributed storage for handling large-scale data processing. Scalability, fault tolerance, and the capacity to manage enormous volumes of data are guaranteed by this configuration. Job orchestration, which controls task scheduling and execution as well as the intermediate outputs produced while processing, uses Spark. By combining HDFS and Spark, the system's speed is maximized and useful features may be effectively extracted from text and audio data.

The two main parts of the system's feature extraction pipeline are text and audio data processing. The Librosa package, which offers tools for in-depth audio analysis, is used by the system to extract several aspects from audio data, including Mel-Frequency Cepstral Coefficients (MFCCs), pitch, and chroma. Tokenization, lemmatization, and cleaning are used as preprocessing techniques for text data, and then a BERT-based model is used for embedding. High-dimensional semantic vectors produced by BERT capture the text's contextual meaning. In order to ensure that data is processed effectively before being transferred back to the HDFS cluster for additional analysis or modeling, the system's architecture also incorporates temporary local storage for managing interim files.

*2) Integration:* Comprehensive feature sets are created for every audio segment by combining the text and audio information into a single, cohesive structure. This procedure entails matching the relevant semantic embeddings from the text data with the retrieved audio features, including MFCCs, chroma, pitch, Mel spectrogram, and Tonnetz. The integrated feature set captures both the prosodic aspects of speech and the more profound semantic implications of the content by combining these acoustic characteristics with the linguistic and contextual information from the text. This combined representation increases the analysis's robustness by allowing machine learning algorithms to identify minor mental health trends and cues. The feature sets can be used effectively in classification models since they are arranged methodically to guarantee compliance with downstream tasks.

*3) Execution:* Two primary scripts that each concentrate on different aspects of the data processing workflow make up the implementation. The entire feature extraction process must be orchestrated, and this is managed by the processdataset.scala script. After determining which patient folders are kept in the HDFS cluster, it downloads the necessary files, launches the extractfeatures.py script to extract the features, and then uploads the data that has been processed back to HDFS. However, the extraction of both text and audio features is the focus of the extractfeatures.py script. It makes use of Transformers for BERT-based text embedding, Pandas for data manipulation, and Librosa for audio analysis. The script's output is a CSV file with each patient's extracted features, which may be utilized for further modeling and analysis work.

## VI. RESULTS AND DISCUSSION

The precision, recall, F1-score, and support for PTSD identification are displayed in table II. It draws attention to the variations in performance between Class 0 and Class 1.

TABLE II: Metrics for PTSD Classification

| Class | Precision | Recall | F1-Score | Support |
|---|---|---|---|---|
| 0 | 0.93 | 0.96 | 0.95 | 344 |
| 1 | 0.94 | 0.90 | 0.92 | 239 |

TABLE III: Class-Wise Metrics for PTSD Detection

| Metric | Precision | Recall | F1-Score | Support |
|---|---|---|---|---|
| Accuracy | - | - | 0.93 | 583 |
| Macro Avg | 0.94 | 0.93 | 0.93 | 583 |
| Weighted Avg | 0.93 | 0.93 | 0.93 | 583 |

Test Accuracy: 0.93

Overall metrics for PTSD identification, such as accuracy, weighted average values for precision, recall, and F1-score, and structural average, are displayed in table -III.

*Metrics for Depression*

TABLE IV: Metrics for Depression Classification

| Class | Precision | Recall | F1-Score | Support |
|---|---|---|---|---|
| 0 | 1.00 | 0.84 | 0.91 | 371 |
| 1 | 0.78 | 1.00 | 0.88 | 212 |

The depression detection classification metrics, including precision, recall, F1-score, and support for each class, are displayed in table IV

TABLE V: Class-Wise Metrics for Depression Detection

| Metric | Precision | Recall | F1-Score | Support |
|---|---|---|---|---|
| Accuracy | - | - | 0.90 | 583 |
| Macro Avg | 0.89 | 0.92 | 0.90 | 583 |
| Weighted Avg | 0.92 | 0.90 | 0.90 | 583 |

Test Accuracy: 0.92

Overall metrics for PTSD identification, such as accuracy, weighted average values for precision, recall, and F1-score, and structural average, are displayed in table -V and Missing values are showed with (-) to indicate unavailable metrics.

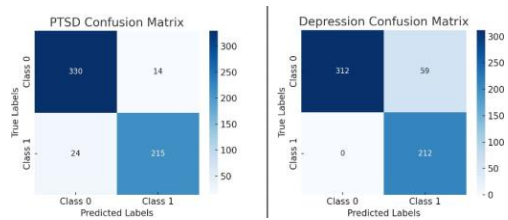

Fig. 3: Confusion Matrix for PTSD and Depression

Figure 4 shows the PTSD and Depression Classification Metrics, a bar chart that depicts the Precision, Recall, and F1-Score for each class, respectively, while Figure 3 illustrates the Confusion Matrix for PTSD and Depression.

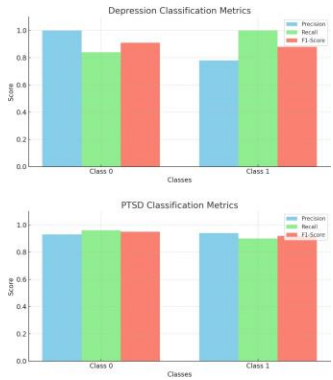

Fig. 4: PTSD and Depression Classification Metrics

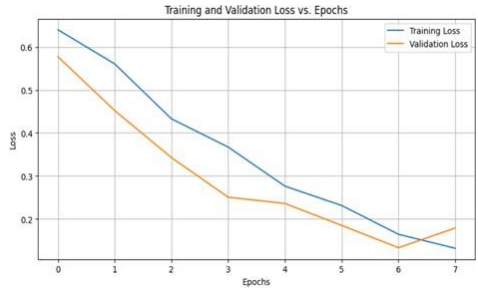

Fig. 5: Loss vs Epoch for Depression

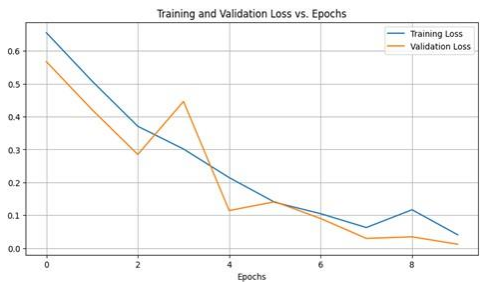

Fig. 6: Loss vs Epoch for PTSD

Loss vs. Epoch for Depression and PTSD are depicted in the following figures, Fig. 5 and Fig. 6, respectively.

A comparison of classification models is shown in the table VI that follows. It draws attention to how well various models perform in relation to important evaluation metrics and and Missing values are showed with (-) to indicate unavailable metrics.

TABLE VI: Comparison Table of Models for Classification

| S.No. | Model | Acc. | Prec. | Rec. | F1 |
|---|---|---|---|---|---|
| 1 | 1D CNN & BiLSTM [12] | - | - | 0.92 | 0.85 |
| 2 | SVM & BiLSTM [13] | - | 0.74 | 0.66 | 0.69 |
| 3 | multi-DDAE, PV [15] | - | - | - | 0.72 |
| 4 | CNN, SVM, KNN, RF, LR [16] | 0.77 | 0.78 | 0.98 | 0.87 |
| 5 | 3D-CBHGA, RF [7] | 0.77 | 0.63 | - | 0.64 |
| 6 | AMST + BERT CNN [3] | 0.86 | 0.88 | 0.92 | 0.90 |
| 7 | LSTM, BiLSTM (Proposed) | - | - | - | 0.92 |

A web application built with Flask was created to predict PTSD and depression by utilizing multimodal inputs, such as text and audio files. The system combines text and audio features, using pre-trained models (depressed.h5, ptsd.h5) for reliable predictions. Librosa processes the audio, while a BERT tokenizer processes the text. JSON answers are used to communicate the results.

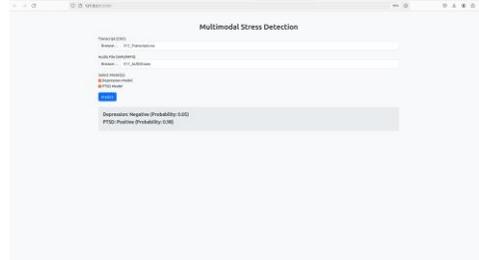

Fig. 7: Screenshot of Web-Interface of Predicted Output

Figure 7 shows a screenshot of the website interface that shows how users can view predictions and upload files. After analysis, temporary files are safely managed and deleted to protect data privacy.

## VII. CONCLUSION

Leveraging BiLSTM and LSTM architectures to integrate temporal information from text and audio data, this study concludes by demonstrating the efficacy of multimodal deep learning models in detecting mental health problems including depression and PTSD. The method shows excellent classification accuracy and highlights how prosodic and semantic cues can be combined to provide reliable mental health diagnostics. The system will be improved in future research by adding more modalities like gestures, facial expressions, and physiological signals. This could increase prediction accuracy and offer a more comprehensive understanding of mental health issues. The system's usability and accessibility in actual clinical settings may be improved by extending its cross-lingual capabilities and integrating it with healthcare systems. However, to ensure fairness, dependability, and user trust in automated systems, it is crucial to address constraints including dataset imbalance, the intrinsic complexity of mental health diagnosis, and ethical concerns surrounding privacy.

## VII. ACKNOWLEDGEMENT

This research supported by Council of Science and Technology (CSTUP), Sanction letter no– CST/D-71, (Project ID- 3965) Authors are thankful to CSTUP for providing us required fund for the research. Authors are also thankful to the authorities of Indian Institute of Information Technology, Allahabad at Prayagraj, for providing us infrastructure and necessary support.